# TOWARD ORGANIC COMPUTING APPROACH FOR CYBERNETIC RESPONSIVE ENVIRONMENT

Clement Duhart[1,2] and Cyrille Bertelle[2]

[1]ECE Paris, LACSC, France
[2]Normandie Univ, LITIS, FR CNRS 3638, Le Havre University, France

*ABSTRACT*.

*The developpment of the Internet of Things (IoT) concept revives Responsive Environments (RE) technologies. Nowadays, the idea of a permanent connection between physical and digital world is technologically possible. The capillar Internet relates to the Internet extension into daily appliances such as they become actors of Internet like any hu-man. The parallel development of Machine-to-Machine communications and Arti cial Intelligence (AI) technics start a new area of cybernetic. This paper presents an approach for Cybernetic Organism (Cyborg) for RE based on Organic Computing (OC). In such approach, each appli-ance is a part of an autonomic system in order to control a physical environment. The underlying idea is that such systems must have self-x properties in order to adapt their behavior to external disturbances with a high-degree of autonomy.*

## 1 INTRODUCTION

Since the 90's, the developpment of enabling technologies for Internet of Things (IoT) has received a lof of interests from di erent research communities. The identi cation of people and things has permitted their representation into digi-tal world through Radio Frequency IDenti cation (RFID) technologies. This has permitted a lot of applications to be developed for logistic traceability and ac-cess control in many domains such as transportation, industrial or building. The growing use of such devices for Smart-Structure has required Wireless Sensor and Actor Networks (WSAN) in order to monitor and manage them. A lot of issues is appeared to establish a heterogeneous network infrastructure which must be energy e cient with several manufacturers and services. Hence the middlewares have received a particular attention in order to abstract such issues to applica-tions of future services. Nowadays, there are well-de ned standard protocols like IPv6 LoW Power Wireless Area Networks (6LoWPAN) with a lot of middle-ware solutions. The extension of Internet into the daily environments of human is coming to regulate shared resources like energy or to improve accessibility, protection and assistance for persons. The term of Ambient Intelligence (AmI) groups the whole of research activities to design control systems for such appli-cation. However their information architectures stay an open question regarding the management of these distributed systems in large scale networks

In this paper, an approach based on Organic Computing (OC) is investigated in order to design autonomic systems which must be autonomous to achieve their tasks and adaptive to address runtime issues produced by unpredictable disturbances. The Section 2 presents related works regarding AmI by introducing the recent interests for OC. Then the Section 3 presents the proposed model for Organic Ambient Intelligence (OrgAmI) with the rst implementations of function components in Section 4. Finally the conclusion with perspectives of this proposal are presented in Section 5. This contribution synthesises several papers already published or under submission in order to present a general overview of the proposed model which is implemented in EMMA framework.

## 2. RELATED WORKS

### 2.1 Ambient Intelligence

The term of AmI is refer to application of Arti cial Intelligence (AI) on Re-sponsive Environments (RE). A lot of approaches for AmI are presented in the survey of Sadri et al. [13]. Two main categories of models can be proposed:

Rule-based and Context Awareness. A Rule-based model de nes a set of rules to apply actions according to a system state such as the Event-Condition-Action (ECA) model of Augusto et al. [3,2] in a Smart Home application. The set of required rules to manage complex situation is very di cult to de ne. Hence Ac-tion Planning focuses on the determination of the best path of actions according to current system state such as presented in Simpson et al [15]. Decision Tree (DT) is a sub-model to consider the possibility of deviations from the expected result to the nal environment state because of hidden variables. Stankovski et al. [16] presents an application on Smart Home with successfull results. However this model-based approach requires a full knowledge of possible states of the environment which are not always easily to model due to discretization and ac-curancy issues. Even if theFuzzy Logic is a model to relax the speci cation of rule description such as investigated in Doctor et al [5], the consistency of large rule set is a strong issue. Reasoning Engines focus on general context of the system instead of particular variables to determine the system state. Magerkurth et al. [10] presents an architecture based on ontologies which are meta-descriptions of the system. The states are not only depending of independent variables but also their relation. The system infers the real system state according to the match-ings between ontology patterns and the internal state representation. However an ontology stay a model which must be de ned. Aztiria et al. [4] presents a survey of learning technics in order to build empirically the system model. On one hand the model is extracted online from system runtime and on the other hand it can be improved continously. However learning process requires a lot of data which are not always easily available to design operational systems.

Finally, there are a lot of AI technics to design decision systems, however the control of RE stays an issue. The managed environment is an open and dynamic environment which is inherently a Complex System (CS). The new approach of Multi Agent System (MAS) considers each appliance such as an autonomous agent implemented with previsouly presented AI technics. The global system behavior is produced by the self-organization of agent interactions such as in Da Silva et al. [14]. Andrushevich et al. [1] use the MAPE-K model in order to design supervisor for such MAS application. It is composed of four modules: Monitor-ing, Analysis, Planning and Execution which exchange information through a common knowledge space. This control loop is a common model to design au-tonomic system e.g. systems with self-x properties to manage CS.

## 2.2 Organic Computing

M•uller-Schloer et al. present in their book [11] a compilation of papers working on OC. This new paradigm is motivated by the necessity to de ne a framework in order to control self-organization process in large and complex networks. A lot of contributions in Multi Agent System (MAS) proposes frameworks with self-x properties, however there is no well-de ned and common de nitions. In this book, the contributors propose a framework to de ne mathematically the terms of au-tonomy, organization, adaptivity, robustness and others with their relations. OC framework focuses on methodology to develop self-organized MAS with trustwor-thy responses. Stegh•ofer et al. [17] discuss about the challenges and perspectives of such systems. Trustworthy systems provide a correct response independently to internal or external disturbances by a self-adaptation of its internal param-eters. The idea of autonomous system is concerning in the reduction of control parameters for the system by a supervisor. They are self-established by the whole system which is composed of a System under Observation and Control (SuOC) and its supervisor. Kasinger et al. [12] present the generic Observer-Controller model for OC systems. The component observer evaluates the current situation of the SuOC according to an observation model selected by expert. This model con gures the observer engine by selecting the proper analyzers and event pre-dictors to provide a context information to the controller. This last one evaluates the possible actions according to the constraints and selects the best one accord-ing to its history database. The constraint set is updated by a simulation engine in order to adapt human requirements to the SuOC.

Finally the computational models of AmI are incoporated into OC architec-tures to built autonomic systems which should be able to evolve according to their environments. The proposed contributions are motivated to design Cyber-netic Organism (Cyborg) in line with the observation that evolution has already produced components required in autonomic system. Hence this work focuses on the basic mechanisms used in biological metabolism in order to propose a bio-inspired framework for Organic Ambient Intelligence (OrgAmI).

## 3 ORGANIC AMBIENT INTELLIGENCE

In following document, the term of Organic Ambient Intelligence (OrgAmI) is refering to a decision system based on an arti cial metabolism between the di er-ent appliances of the System under Observation and Control (SuOC). They are considered such as cells which communicate through endo- and exo-interactions in order to satisfy system goals under its constraints. The general architecture overview in Figure 1 is composed of three main levels corresponding to the MAPE-K model for autonomic system. Each appliance is a System on Chip (SoC) which communicates through a Wireless Sensor and Actor Networks (WSAN) in order to share its specialized services. Then these cellular services exchange information thanks to reactive rules based on Event-Condition-Action (ECA) model executed on a Resource Oriented Architecture (ROA) middleware. An OrgAmI function is an interaction graph of rule based agents which produce computation ows between the resources of cellular services. Hence supervisors are responsible to analysis OrgAmI behavior to learn interaction patterns in order to control the metabolism of the SuOC by updating reactive rules. An OrgAmI function, named Service Choreography (SC), is a distributed control loop between sensor and actuator cellular services.

In practice, the daily use of the managed environment by human modi es the system metabolism. Then Observer-Controller supervisors extract reactive rule patterns in order to built or update knowledge. Hence this knownledge is instanciated according to goals and constraints of the SuOC in order to deploy or update online the SC. This knwoledge is built by symbolic inference from a statistical learning based on Arti cial Neural Network (ANN) in order to be shared and analyzed by human experts.

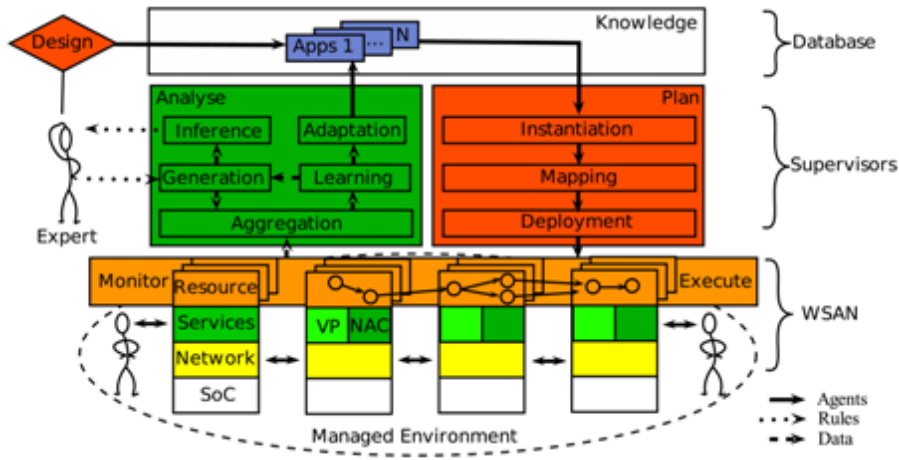

Fig. 1: General overview of OrgAmI architecture.

## 3.1 Cellular Resource Middleware

The Cellular Resource Middleware (CRM) abstracts hardware and network im-plementations to the metabolism layer such as presented in [9]. A WSAN is composed of tiny platforms which execute hard-coded services such as actuator and sensor drivers, system components or advanced processing. This specialized cellular services exchange information through a resource interface which man-age interactions with the metabolic environment like the membrane of biological cells. For example, an output resource connected to an actuator is driven through a distributed computation ow of cellular interactions over the SuOC when an event appears on an input resource connected to a sensor. A Service Choreog-raphy (SC) is a graph of such service interconnections. The ROA middleware illustrated in Figure 2 is composed of three basic cellular services used to model such distributed computation ows:

  { Local Resources (/L/) are memory units to store logical or numerical values.

  { System Resources (/S/) are input-output interfaces of cellular services.

  { Agent Resources (/A/) are script les to model interaction mechanism in which an agent updates a target resource by a preprocessed payload accord-ing to current state of node resources.

The Figure 3 illustrates a computation ow in which agents update remote resources which trigger other agents like a domino e ect. In the metabolic point of view, the activation of an agent is depending of the internal state of its cell in order to emit an interaction to another cells which produce an interaction computation ow over the SuOC.

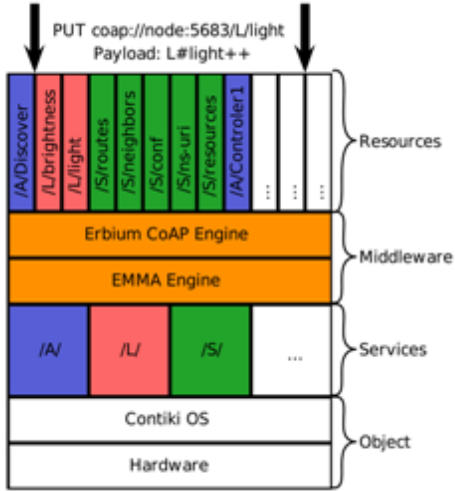
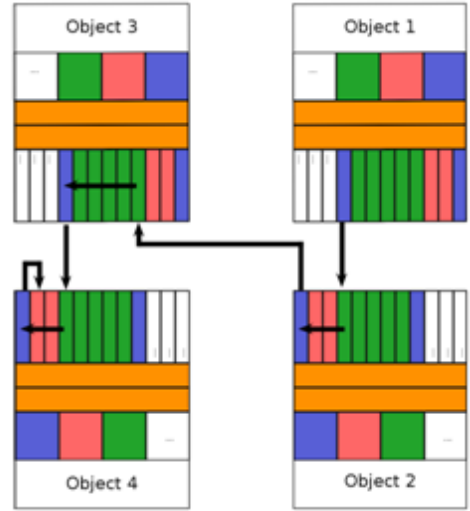

Fig. 2: Cellular middleware.    Fig. 3: Metabolic computation flow.

## 3.2 Formal Agent Model

In Definition 1, an agent is an Event-Condition-Action (ECA) rule in order to emit an interaction to a resource of a cellular service.

Definition 1. An agent a on cell n is a reactive rule to update target resource $y \in Y_m$ of the cell m by the post-processed function $POST_a^y(X_n; X_m)$ regarding of the internal resource states of $X_n$ and $X_m$ with the interaction operation of creation, deletion or update when the activation function $PRE_a(X_n)$ is true.

$$If\ PRE_a(X_n) : \forall y \in Y_m, y \xleftarrow[operation]{} POST_a^y(X_n, X_m)\ with\ Y_m \subset X_m \quad (1)$$

A Service Choreography (SC) is a set of interconnected services in order to control actuators according to the SuOC context composed of their state and those of the sensors. The release of a new sensor value in a resource produces an event on the cell which can activate others agents such as in Definition 2.

Definition 2. A metabolic computation ow $f \in F$ is an activation chain of agents updating the resources of cellular services along the diffusion path of pro-duced events on corresponding cells.

The proposed model to design SC is based on an adapted Petri Network in which agents are represented by transitions and resources by places such as il-lustrated in Figure 4. Hence the metabolic processes are analyzed to validate their termination, input-output range and other properties by classical algo-rithms found in Petri Network literature. In addition, this model is completed by associating to each transition a place such as agents are also resources which can be modified. This adaption has strong implications such as the SC becomes dynamic with a Petri Network which evolves during its execution. In such sit-uation, the metabolism layer is a Complex System (CS) which cannot have a predictable behavior by an analytic approach such as it is open to external events and has self-rewriting capacity. In Organic Computing (OC) model, this issue is addressed by an offine simulation to evaluate SuOC configurations.

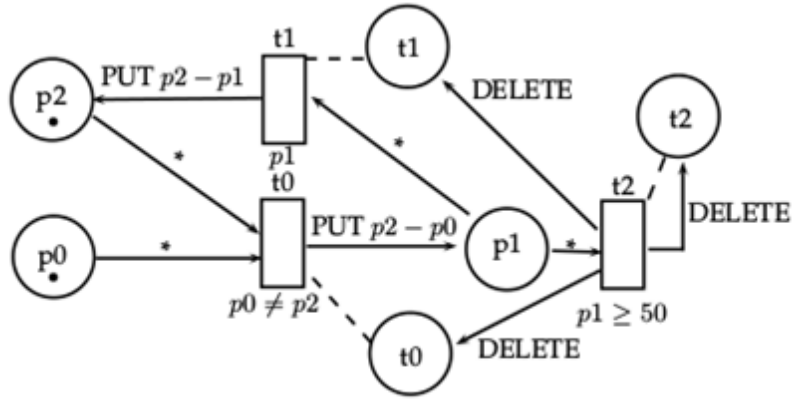

Fig. 4: This SC computes p1(t) = p0(t 1) p0(t) through agent t0 and t1. If the value 50 is reached, the agent t2 deletes the metabolic process.

**3.3 Metabolic Mapping Deployment**

The metabolic mapping detailed in [6] consists in deploying the different re-sources over the WSAN according to cellular services and requirements of the OrgAmI. This process is composed of two steps: the first one is performed by a preprocessor which determines the SC to deploy according to a WSAN target. Based on an intermediate graph of instanciation, a Pseudo Boolean Optimization (PBO) problem is formulated to compute the best resource mapping in order to minimize network communication load.

Finally, there are several available deployment strategies based on self-writing cellular property. The agents have the ability to create new ones. Hence a gen-eral agent can contain other agent to deploy which can themselves contains other agents like a Matroska. Such composed agent allows the supervisor to delegate the deployment process to any cell. Otherwise it can generate a general agent which is responsible to deploy every resources by accrossing all cells. In such ap-proaches, the SC called Residual Network Agent (RNA) are deployed by another SC named Dynamic Network Agent (DNA). Hence the cells are programmed by a DNA process which deployed a RNA behavior.

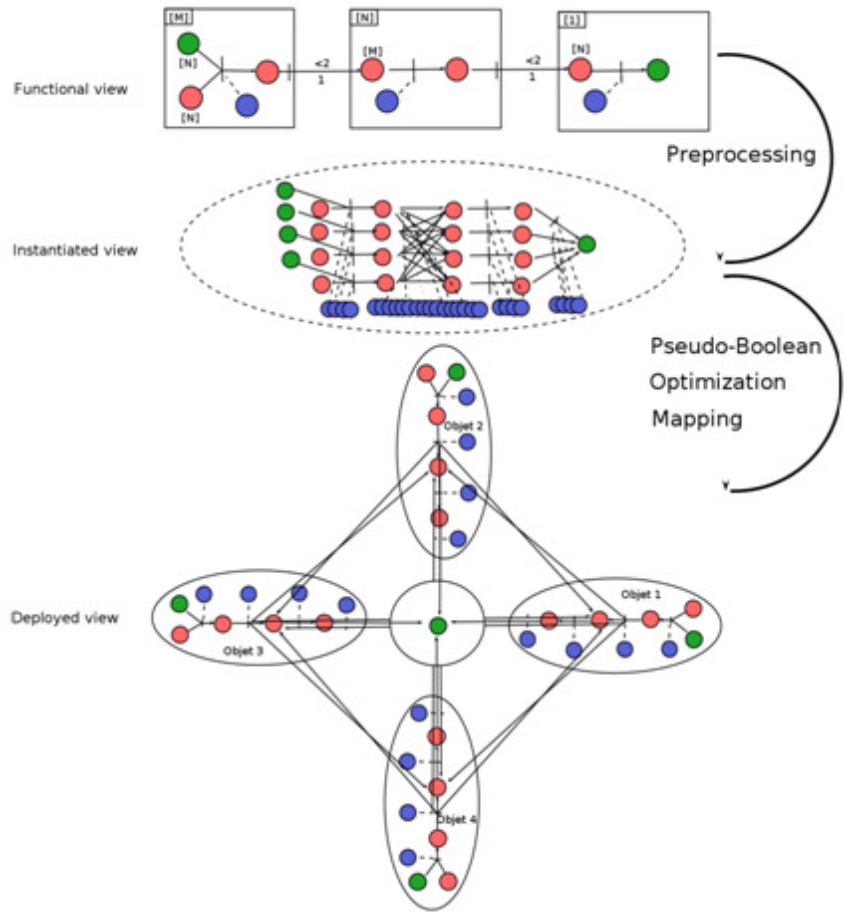

Fig. 5: Scheme overview of the di erent steps for mapping and deployment of SC stored in knowledge database onto Cellular Resource Middleware (CRM).

## 4. APPLICATION: CEREBRAL CORTEX COMPONENTS

This Section presents two applications of presented OrgAmI framework in order to model Cerebral Cortex (CC) components distributed over WSAN. In bio-logical brain, the CC contains major cognitive functions of sensing, acting and association. Following work presents rst implemented components of a toolbox for the design of future cyberbrains applied on Ambient Intelligence (AmI). The Motor Cortex (MC) is responsible of the planni cation, control and execution of movements which relates to actuator control. The Anterior Cingulate Cortex (ACC) ensures behavior consistency in case of cognitive function con icts.

### 4.1. Arti cial Neural Controller

The Motor Cortex (MC) is a brain region reponsible of the planning and the execution of movements. The learning process is based notably on neural plas-ticity in order to establish synaptic connections between interneurons, Betz cells and others. Similarly in Arti cial Neural Network (ANN), the learning process determines such synaptic weights in order to control output neurons.

The Arti cial Neural Controller (ANC) presented in [7] has been studied in order to control actuators according to the system context composed of their initial states and those of sensors. The proposed methodology to design ANC has permitted to conclude that it is preferred to use several specialized ANC instead of general one. Similarly, MC have specialized areas. Hence an ANC is composed of several ANN representing di erent behaviors. They are selected according to their relevance to perform the task such as illustrated in Figure 7.

The Figure 6 presents its integration into the CRM. Even if the neurons can be directly modelled by CRM with an activation function based on agent and weights on L resources, the neural behaviors are stored into resources of a specialized cellular service for execution e ciency reason. Their training is performed remotely by collecting data when there is no relevant behavior. Finally the ANC service is connected to actuator and sensor resources by metabolic SC.

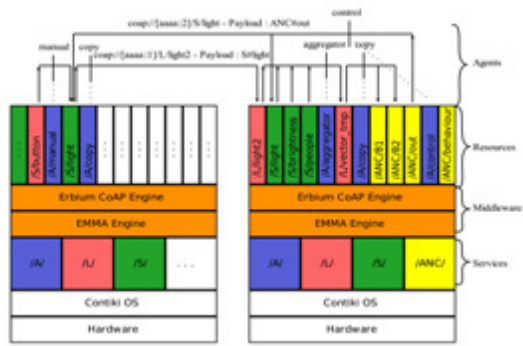
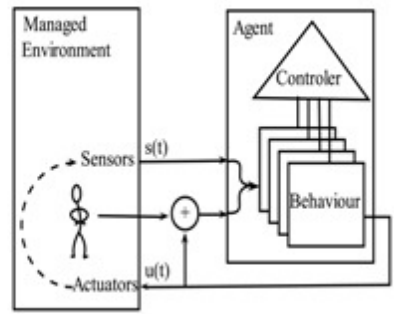

Fig. 6: ANC middleware integration.	Fig. 7: ANC overview architecture.

The Figure 8 presents a runtime example of an ANC with one initial ANN behavior. The plot (a) shows the input-output signal composed of  ve resources. The ANC should predict the correct next states of resources in order to control them before human acting manualy on actuators. Hence sequential plans are learnt state by state. The plot (b) presents prediction errors of the ANN behav-iors in order to decide between the selection of the best one or the learning of a new one. The behaviors are ordered in plot (c) according to their successive prediction errors during runtime in order to select the best one in plot (d). At initialization, there is only one behavior and it is not able to predict properly the next states. Hence the ANC collects data in order to learn a new ANN. When this signal appears a second time, this behavior is correctly selected.

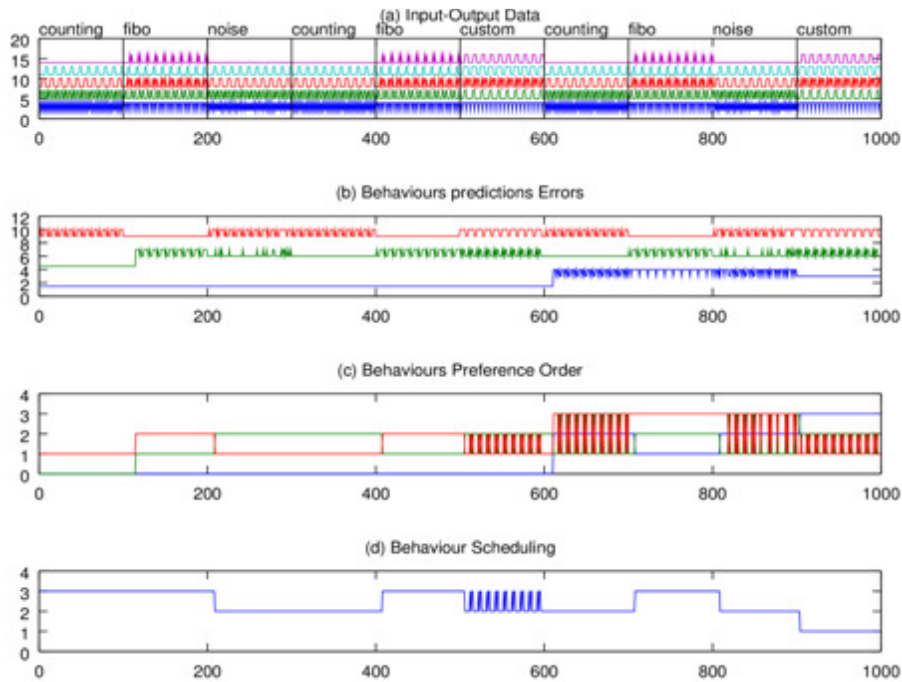

Fig. 8: ANC runtime example in which three different behaviors are learnt before being scheduled in order to control the input-output corresponding signals.

Finally a methodology is proposed to generate statistical data set from a model of Decision Tree (DT) in order to predefined behaviors. They are up-dated by the mixing of generated and collected data from runtime SuOC. The human actions on actuators refine the behaviors during their learning process by correcting the ANC decisions. Then DT are extracted from learnt behavior using A-Priori algorithm in order to expand the knownledge database.

**4.2 Distributed Voting Procedure**

The Anterior Cingulate Cortex (ACC) has an important function in arbitration of cognitive function conflicts. The Stroop task proposed by Pardo et al. in 1990 produces its activation when a conflict appears during the reading of the word RED written in blue during a serial test. Its primary function of conflict monitoring stimulates other brain areas in order to resolve it.

The Voting Procedures (VP) presented in [8] has been developed to address such issues in cyberbrain with an implementation on EMMA framework. In this paper, the SC are abstracted by a set of decision-makers which aggregate their preferences in conflicting in order to take a common distributed and consistent decision. It is composed of an aggregation process illustrated in Figure 10 and a decision step. The aggregation is based on Network Average Consensus (NAC) algorithm which aggregates the utility values of agent preferences towards an averaged preference order in a fully distributed way. The NAC have very inter-esting properties of fault-tolerance regarding switching network topology, packet lost and delayed transmission which are inherited by the VP. Moreover this al-gorithm is more efficient in terms of resolution time on Small-world topologies which can be also found in biological brain. Hence, the decision step evaluates the accuracy of the aggregation in order to re ne it to guarantee that there is not ambiguity between decision-makers.

The Figure 9 illustrates the integration of VP algorithm on CRM. It is fully implemented by a SC. There are NACx agents which are reponsible of the x aggregations of preference values. An aggregation agent decides to stop it when the accurancy is enough in order to allow a selection agent to take a consistent decision. Otherwise the aggegation is repeated until reaching the required accu-rancy. The Figure 10 illustrates the aggregation process of ten decision-makers which have to select a parameter among ten. This process requires 50 iterations according to the initial preference orders and the WSAN topology.

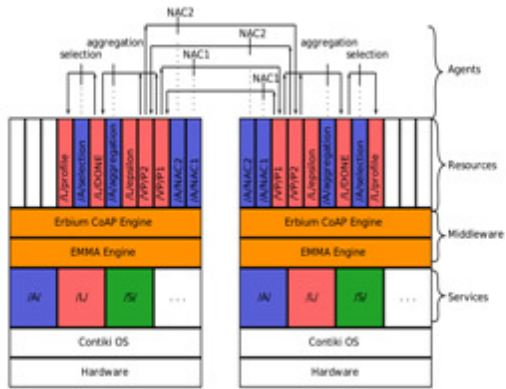 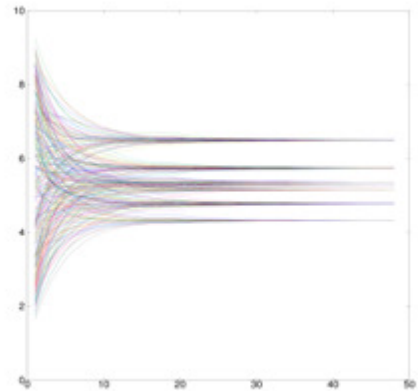

Fig. 9: VP middleware integration      Fig. 10: VP runtime example

## 5. CONCLUSION

This paper is the rst presentation of the underlying Organic Ambient Intel-ligence (OrgAmI) model in continuity of the software suite developpment of

Environment Monitoring and Management Agent (EMMA) . Its di erent com-ponents have already been published or are under submission in the di erent technological and theorical disciplines of Wireless Sensor and Actor Networks (WSAN), Multi Agent System (MAS) and Organic Computing (OC).

This bio-inspired framework focuses on the possibility to design fully dis-tributed Ambient Intelligence (AmI) functions at the elementary level of metabolic systems. It is assumed that natural systems are strongly autonomous and adap-tive thanks to their design by elementary assemblies which o ers them large possibilities of system recon gurations. Hence the proposed OrgAmI is a Sys-tem under Observation and Control (SuOC) based on an interaction metabolic model enabling self-x properties under the control of high level supervisors. Two rst components are implemented thanks to this bio-inspired framework in order to start investigation on the possibility to design distributed cyberbrain archi-tecture for Responsive Environments (RE).

The development of an operational cyberbrain based on OrgAmI is still a long-term perspective. On one hand, a lot of neural functions found in biological brain should be implemented before to be able to address cognitive aspect of AmI. And on the other hand, the control of such fully distributed system in which computations are operated at an interaction level required advanced analysis and planning tools to manage such autonomic system.

The short-term development will address a use case of energy management ar-bitration in Smart Home application. A set of Arti cial Neural Controller (ANC) controls di erent appliances which negotiates together the energy thanks to the proposed Voting Procedures (VP). Hence the RE will learn to automatically con-trol the appliances according to their uses by human and an implicit scheduling based on energy availability during one-day period.

## ACKNOWLEDGEMENTS

The authors thank M. Mardegan and M. Sauvage for their contributions during the developpement and implementation of EMMA framework.